# Robotic Non-Destructive Testing of Manmade Structures: A Review of the Literature

J.D. Zhao

**Abstract** — This literature review investigates how robots can be used for the maintenance of manmade structures, such as pipes, reinforced concrete decks, and space stations as a sampling of the broad spectrum of robot non-destructive testing (NDT) applications. Robotic NDT can be used to find plaque in pipes, corrosion in steel buildings, and impact damage in space stations, which would normally be invisible to the eye. After inspection, the inspected material is preserved in its original condition. This paper's structure is as follows: first, the definition of NDT is elaborated upon with the discussion of specific methods that will be used in the inspection of structures mentioned above. Second, an explanation follows on why robots are suited to inspection, specifically focusing on robots' advantages over humans. Third, three real world examples notify the reader on current progress in robot NDT. Lastly, a summary of robot problems serves as a reminder that testing and development must continue for robot NDT to become mainstream.

— — — — — — — — — ◆ — — — — — — — — —

## 1 INTRODUCTION

The convenience of our modern lifestyle is made possible by manufactured products. Everything that we use--from phones to cars--was once made in a factory, pieced together on the assembly line. And from time to time, mistakes happen. A phone may be produced which has a faulty screen, or a car whose engine won't start. It's clear that these defective products shouldn't be in the market, because they do not create profit for the manufacturer and because they pose a safety hazard for the user. For example, what if the faulty screen exploded, or if the car engine suddenly failed in the middle of the highway? Angry consumers would demand a refund from the manufacturer. Because of this risk, manufacturers have implemented NDT that weeds out defective products. According to Carmelo Mineo, an engineer from the University of Strathclyde, the same processes are used in the diagnosis of nationwide infrastructure (2012).

Seventy years ago, during America's rise as a global power, a nationwide, well designed infrastructure network was built. But today, it is deteriorating. The surfaces of pipes in nuclear power plants have become eroded because of continuous water flow, threatening a catastrophic leakage. Reinforced concrete buildings, consisting of concrete shells supported by steel skeletons, have rusted from rain. Even the pinnacle of engineering, the International Space Station, has slowly become damaged due to space debris, as two Jet Propulsion Laboratory engineers note (Volpe and Balaram 1994). Yet the deterioration is so widespread that human inspectors have difficulty pinpointing and making repairs. Today, a new solution has the potential to remedy this, called robotic non-destructive testing. Robotic NDT offers exciting new possibilities in freeing up engineers' time and saving money, and has implications for halting the decline of American infrastructure.

## 2 WHAT IS NON-DESTRUCTIVE TESTING?

Non-destructive testing (NDT) "Is a wide group of analysis techniques used in science and industry to evaluate the properties of a material, component or system without causing damage," according to Louis Cartz, an engineer from Marquette (Cartz 1995). It is an umbrella term for techniques such as piezoelectric, electrochemical, and machine-vision inspection, described below. Manufacturers highly value NDT because it does not affect the material that is being tested; manufactured products such as phones and car engines can be harmlessly checked for defects, ensuring total safety. NDT is also applied in aerospace and civil engineering, checking the extent of damage on space stations and calculating the lifespan of buildings (Volpe 1994). Because of its widespread application to so many industries, NDT is a technique that has already caused great benefit and can be improved upon by integration with robots. The following are NDT methods ideal for use with robots.

### 2.1 Piezoelectric Inspection

The word piezoelectricity comes from two Greek words: piezo, meaning to squeeze or press, and electric, meaning amber, which when rubbed with fur becomes positively charged (Holler 2007). Piezoelectricity is a phenomenon present in solid materials such as crystals and certain ceramics. When a crystal such as quartz are squeezed, it accumulates electric charge. The deformation to the crystal caused by the squeezing is so tiny that it cannot be observed by the naked eye, yet it produces a measurable electric field. Piezoelectric sensors consisting of crystal elements and wiring, are suitable for failsafe robot inspection. The Indian Institute of Technology has designed a cylindrical robot with rotating piezoelectric sensors, driven wirelessly inside an old pipe, to make a voltage "map" of the pipe's interior, allowing repairmen to know where the cracks and bumps are (Kentarou, Yixiang and Rupesh 2010)..

### 2.2 Electrochemical Inspection

According to Corrosionpedia, the definition of electrochemical corrosion potential is "The voltage difference between a metal immersed in a given environment and an appropriate standard reference electrode." Essentially, the electrical properties of the test metal and a reference metal are compared using a voltmeter. The differences in the properties compared determines the level of corrosion in the test metal, and whether or not a repair will be made. For example, according to Kenji Reichling, a civil engineer from Aachen University, Germany, "there is a high risk of chloride induced reinforcement corrosion" on the steel reinforcement in concrete parking build-ings" (2009). To accelerate the inspection of the steel rebar, an autonomous robot, BETOSCAN, was developed. Equipped with electrochemical sensors, it is able to simultaneously navigate over flat surfaces and take measurements, allowing the condition of a parking deck to be determined in a day (Reichling 2009).

### 2.3 Machine Vision Inspection

Video inspection, one form of visual inspection, is a common and affordable technique. In this method, the surface of the structure, whether it be a bridge or metal plate, is slowly scanned by some video device. The video can then be passed on to either a human or a computer. If a human is involved, he or she will simply look for cracks. Video inspection is advantageous because it allows humans to inspect areas which would pose hazards to them, such as in outer space (Mineo 2014). The video can also be passed onto a computer, where the computer will take snapshots and compare it with a reference image. Through an image differentiation program, any changes, such as damage, can be spotted. A video inspection system, consisting of visible light, ultraviolet, and infrared cameras mounted on a robotic arm has been developed for detecting meteorite damage on the International Space Station (Volpe 1994).

## 3 WHY ARE ROBOTS GOOD CANDIDATES FOR PERFORMING NON-DESTRUCTIVE TESTING?

Despite NDT's importance in ensuring the quality of products and structures, current NDT techniques are dominated by manual scanning of a material's surface. According to Carmelo Mineo, "manual scanning requires trained technicians and results in a very slow inspection process for large samples" (2014). In addition to the slow speed, trained technicians are expensive because they have undergone specialized training. Although NDT robots do initially cost more than human labor, robots do not tire, do not get bored, can work in hazardous conditions, and cannot go on strike. If robots were introduced for NDT, productivity per person per hour would be increased as a single human technician could manage multiple robots. In addition, statistics show that there are fewer young people entering the NDT profession, and providing robots could compensate for this dearth. Lastly, robots now have become more sophisticated and accurate, with the ability to repeat positions to the accuracy of one tenth of a micrometer (1 / 10,000,000 m). Because robots have become more precise and less costly, as well as the capability for one human operator to control multiple robots, the current NDT bottleneck can be alleviated with robots.

### 3.1 Costs of human labor

While the loss of a human life is a terrible tragedy, the loss of a robot can be replaced by purchasing another one. Christos Emmanouilidis, PhD in electrical engineering and expert in artificial intelligence, advocates robot NDT for testing steel welds because of high temperatures and fumes. Indeed, "robotic solutions are increasingly replacing humans in carrying out risky non-destructive testing (NDT) tasks. The advantages are profound as automated inspection can reduce risks posed to human operators, costs involved in laborious manual inspections and errors introduced by the operators" (Emmanouilidis 2004). In addition, humans are not willing to perform NDT in places where there is a danger of structural failure, such as inside an old masonry building after an earthquake. The lightness and maneuverability of robots could eliminate the riskiness involved in inspection and cause as little disturbance to the structure as possible. "There are numerous applications of climbing robots and autonomous miniature robotic vehicles for use in NDT these being suited to inspection of large structures where access by human operators may be hazardous or limited," according to Mineo (2012). Using robots to perform NDT in dangerous places will mitigate loss of life when an accident inevitably occurs.

### 3.2 The Efficiency of Robots

Although humans are better than robots at tasks that require novelty, humans easily get bored when confronted with a rote task, which can lead to carelessness. At simple tasks common in NDT, robots are far more efficient. Kenji Reichling, an academic at the Institute of Building Materials Research, Germany, and several other researchers developed a robot, named BETOSCAN, that could autonomously navigate level concrete surfaces common in tiered parking garages. The problem they encountered was that parking structures are made up of reinforced concrete; the steel is subject to corrosion by chloride and deicing salts. Conventional NDT using handheld electrodes to determine which areas were most corroded took a long time and was expensive, requiring people to walk to map out every square meter of the building. In lieu of any other method, shorter, less rigorous inspections were carried out, as described below.

*"To evaluate the condition of such structures adequately extensive investigations are usually necessary. Frequently the costs of a complete investigation exceed the available funding, so that only abbreviated investigation programs are carried out. The results of such programs are not usually sufficient as a basis for an adequate design of repair and protection measures." (Reichling 2009).*

However, once BETOSCAN was implemented, a whole parking garage could be scanned and analyzed in one day without much supervision on the robot, drastically increasing efficiency. In the field of digital radiography, robots are also more efficient than human radiographers. Radiography is a commonly used procedure to search for cracks or flaws in welds for metal parts on the assembly line, the speed of which was hampered by technicians manually positioning a device to scan each part individually. With the creation of robot arms that could perform complex manipulations, NDT radiography could now be delegated to robots. Radiography expert Daniel W. Bosserman explains.

*"By programming a robot to perform the actual labor, a skilled radiographer can increase exposure time, produce more radiographs of consistent quality and avoid the boring, hazardous elements of the job. And, because a radiographer can position and control more than one robot, productivity per man-hour can be multiplied" (Bosserman 2007).*

For example, if four radiographers were once used to scan parts on a single assembly line at constant speed, and, due to a recent upgrade, each radiographer was placed in charge of four robots, each capable of handling a single assembly line, productivity per man-hour would increase sixteen fold. When faced with a simple yet boring task, robots have demonstrated that they are far more efficient than humans.

## 4 FOR WHAT MAN-MADE OBJECTS CAN ROBOTS BE USED TO PERFORM NON-DESTRUCTIVE TESTING?

There are an unlimited number of opportunities for robots to perform non-destructive testing. One of them is the testing of steel plates using an array of electromagnetic and ultrasound sensors. In a 2004 study led by Christos Emmanouilidis, artificial intelligence expert, work was done focusing on the automated inspection of steel plates. Such plates are used extensively during the manufacturing process, but high temperatures and heavy machinery make the inspection process difficult. Emmanouilidis focused on creating a mobile robot, much like a self-driving car, for inspecting steel plates. In another study, a collaborative effort between Waseda University in Japan and the Indian Institute of Technology, researchers developed a robot for the inspection of drainpipes, using a touch sensor to detect flaws (Kentarou 2010). Numerous other applications of robots have been developed to scan concrete parking structures, detect flaws in carbon fiber plates used for commercial aircraft, and for the inspection of space stations. The versatility and robustness of robots make them well suited to performing inspection on all man-made structures.

### 4.1 Steel Plates

Steel is an important material in many industries and is renowned for its tensile strength. Steel plates are used in ships, to store oil, and to build cars and buildings. However, "a pertinent problem in industry is the presence of weld defects or corrosion on the plates. Extensive corrosion or damage in welded plates can have profound environmental or financial consequences, especially in the case of aboveground oil storage tanks" (Emmanouilidis 2004). Because manual inspection is dangerous and slow, Automation Technologies, located in Athens, Greece, started research on an autonomous mobile robot to inspect the plates. The robot utilizes three probes, two ultrasonic, and one electronic. The authors state that "a mobile robot-based automated non-destructive testing system has been developed for the inspection of large steel plates in factory and in-situ. Three different NDT probes are employed, mounted on two probe holders" (Emmanouilidis 2004). The proposed robot is capable of inspecting steel plates both on the assembly line and "in situ," or at the site where the steel plate is actually used. Additionally, "the system under development is based on a mobile robot platform and is designed to carry and deploy ultrasonic and electromagnetic sensors in order to scan plates for corrosion and welds for defects" (Emmanouilidis 2004). Currently, the robot is undergoing testing of its sensors. Sensors are being placed up against different welds, both good and defective, and once the robot completes testing, it will be redesigned to be more durable.

### 4.2 Pipes

Pipes are an important part of city infrastructure, providing clean water and transporting dirty water to wastewater treatment plants. Through normal use, pipes develop chafes and cracks caused by water pressure. These cracks are inside the pipe, and traditionally require a wired robot to go inside the pipe, transmitting video to the human user. Nishijima Kentarou, a researcher at Waseda University, Japan, explains that "many robots to inspect these pipes were developed in the past, but they had a heavy power supply and a signal wire. As a result, these wires caused problems with the movement of the robot. Therefore, in this research, our purpose was to develop a flexible drain inspection robot using a wireless radio communication system" (Kentarou 2010). The proposed robot includes a rotating touch sensor element, developed by the Indian Institute of Technology. Whenever the robot detected an anomaly in the structure of the pipe, such as a depression, the touch sensor recorded it and transmitted the information via radio signal. The antiquated video and wire system was disregarded in favor of a wireless approach. When testing the robot inside a pipe with simulated defects, the authors "confirmed that we could drive the robot by wireless radio communication system in the inside test pipe and collect the image and some signals from the rotating probe" (Kentarou 2010). The touch sensor of the robot used a material called piezo film, and "when the rotating probe touched the defective area, the piezo film could detect the curve and change of the stress" (Kentarou 2010). The robot uses the level of voltage transmitted by the rotating touch sensor to spot defects, as explained: "When a probe touched the defect, voltage showed a substantial decline. Therefore, we believe that four probes can be used to measure problems in inside pipes caused by corrosion, cracking and breakage" (Kentarou 2010). Currently, the robot inspection system is functional, and a research team is working diligently to increase the sensitivity of the robot detection system.

### 4.3 Space Stations

Satellite communication, surveillance, and research is an important facet of our modern world. Satellites provide us with location data, stream video to our televisions, and update us on the state of the world's forests. Many satellites are in Low Earth Orbit, an area that is dotted with objects one centimeter and below in size, referred to as micro-meteorites. In the late 1990s, NASA proposed to launch Space Station Freedom, later converted into the International Space Station. In simulations run in NASA's Long Duration Exposure Facility, responsible for determining the effects on spacecraft while spending long periods of time in space, it was revealed that damage from micro-meteorites would cause astronauts to be perpetually engrossed in repairing the space station, much of that time scouring areas affected by damage. Work was started on finding an efficient way to scan space station for debris. Richard Volpe and J. Balaram, engineers at the Jet Propulsion Laboratory, summarize their findings:

*"For these reasons, NASA sponsored the Remote Surface Inspection Task (RSI), a five-year technology demonstration task at the Jet Propulsion Laboratory, California Institute of Technology (JPL). The project has developed and systematically investigated methods for telerobotic inspection of Space Station Freedom" (Volpe 1994).*

The NDT technique for spotting debris is visual based. It will consist of spotting differences between an image taken earlier when the space station had no damage, the reference image, and the image taken to find damage, the inspection image. As summarized by Volpe: "the approach adopted for on-orbit inspection of space platforms consists of locating and characterizing flaw-induced changes between an earlier reference image and a new inspection image" (Volpe 1994). However, precautions must be taken for the image-comparison software not to come up with false positives due to lighting and other factors. The image-taking hardware is mounted on the end of a robotic arm, known as an end effector, and consists of other sensors such as gas and proximity sensors. The end effector is attached to "a Robotics Research K1207 arm mounted on a translating platform and controlled by a real-time system employing Configuration Control" (Volpe 1994). Processing of the inspection image "provides ambient light compensation, registration correction, and automatic flaw detection" (Volpe 1994). With more satellites being launched each decade, the risk of damage by space debris increases, and the need for outer space NDT increases as well.

## 5 Conclusion

A robot is only as good as its design and programming allow it to be. Robots have no capacity for self-awareness, since they are not sentient, and require a human operator. For scanning robots which are responsible for the detection of surface defects, the scanned material must be in extremely precise orientation to the scanner. According to Mineo, "a significant problem in the field of production line automation is the design of flexible and autonomous robotic systems able to manipulate complex objects. Most current systems depend on complete knowledge of both the shape and position of the parts" (2012). For the piezoelectric inspection of pipes, no computerized analysis tools, other than the naked eye, have been developed to analyze voltage graphs. There are also problems with the most basic of all non-destructive inspection methods using robots--visual inspection--as well. There are the issues of "imaging repeatability, lighting variation, and misdetection of benign changes" (Volpe 1994). Although robotic NDT may hold the key to revitalizing America's aging infrastructure, there are still problems to be solved concerning robot autonomy and reliability.

## References

Bosserman, Daniel W. "The Future of NDT: Radiography Meets Robotics." Quality Magazine. n.p., 27 August 2007. Web. 15 May 2016.

Cartz, Louis. Nondestructive Testing. Materials Park: A S M International. 1995. Print. 26 May 2016.

Emmanouilidis, Christos; Spais, Vasilios and Hrissagis, Kostas. "A mobile robot for automated non-destructive testing of steel plates." ResearchGate. Automation Technologies, January 2004. Web. 15 May 2016.

Kentarou, Nishijima; Yixiang, Sun; Rupesh Kumar, Srivastava; Harutoshi, Ogai and Bishakh, Battacharya. "Advanced pipe inspection robot using rotating probe." WASEDA University. Proceedings of the 15th International Symposium on Artificial Life and Robotics, held 4-6 February 2010, Beppu, Oita. Web. Accessed 15 May 2016.

Mineo, C.; Herbert, D.; Morozov, M. and Pierce, S.G. "Robotic Non-Destructive Inspection." Bindt. In 51st Annual Conference of the British Institute of Non-Destructive Testing 2012 (NDT 2012): Proceedings of a meeting held 11-13 September 2012, Northamptonshire, UK. Web. Accessed 15 May 2016.

Mineo, C.; Pierce, S.G.; Wright, Ben; Nicholson, Pascual Ian; Cooper, Ian. "Robotic Path Planning for Non-Destructive Testing of Complex Shaped Surfaces." TWI-Global. Proceedings of a meeting held 20-25 July 2014, Boise, Idaho, USA. Web. Accessed 15 May 2016.

Reichling, Kenji; Raupach, Michael; Wiggenhauser, Herbert; Stoppel, Markus; Dobmann, Gerd and Kurz, Jochen. "BETOSCAN -- Robot controlled non-destructive diagnosis of reinforced concrete decks." Non-Destructive Technologies. Aedificatio Publications, 3 July 2009. Web. Accessed 15 May 2016.

Volpe, Richard and Balaram, J. "Technology for Robotic Surface Inspection in Space." Jet Propulsion Laboratory: Mobility and Robotic Systems. Proceedings of the AIAA/NASA Conference on Intelligent Robots in Field, Factory, Service, and Space held 21-24 March 1994, Houston, Texas. Web. Accessed 15 May 2016.